\pdfoutput=1
\documentclass[journal]{IEEEtran}
\usepackage[T1]{fontenc}
\usepackage{amsmath,amssymb}
\usepackage{graphicx}
\usepackage{cite}
\usepackage{array,tabularx,booktabs}
\usepackage{url}
\usepackage[hidelinks]{hyperref}
\DeclareRobustCommand{\doi}[1]{\href{https://doi.org/#1}{\nolinkurl{#1}}}

\newcolumntype{P}[1]{>{\raggedright\arraybackslash}p{#1}}
\newcolumntype{L}{>{\raggedright\arraybackslash}X}



\begin{document}

\title{A Survey on Reinforcement Learning Applications in SLAM}
\author{Mohammad~Dehghani~Tezerjani,
        Mohammad~Khoshnazar,
        Mohammadhamed~Tangestanizadeh,
        Arman~Kiani,
        and~Qing~Yang%
\thanks{This article is a revised version of the paper published in
\emph{Journal of Machine Learning and Deep Learning}, vol.~1, no.~1, pp.~20--31, 2024,
doi: \doi{10.64820/AEPJMLDL.11.20.31.122024}.}%
\thanks{M. Dehghani Tezerjani and Q. Yang are with Computer Science and Engineering,
University of North Texas, Denton, USA (e-mail: mike.degany@unt.edu; qing.yang@unt.edu).}%
\thanks{M. Khoshnazar is with the Institute of Artificial Intelligence, University of Bremen,
Bremen, Germany (e-mail: Khoshnam@uni-bremen.de).}%
\thanks{M. Tangestanizadeh is with Computer Science and Engineering, University of California
Santa Cruz, Santa Cruz, USA (e-mail: mtangest@ucsc.edu).}%
\thanks{A. Kiani is with Electrical and Computer Engineering, University of Maine, Maine, USA
(e-mail: arman.kiani@maine.edu).}}

\markboth{}{}
\maketitle

\begin{abstract}
Simultaneous localization and mapping (SLAM) allows a mobile robot or autonomous vehicle to build a map of an unknown environment while estimating its own pose within that map.
Reinforcement learning (RL), in which an agent learns a decision policy from interaction and reward, has been applied to decide how such systems move, explore, and recognize places they have visited before.
This survey reviews the applications of RL in SLAM.
We first distinguish passive SLAM, in which the robot's motion is not chosen by the SLAM system, from active SLAM, in which it is, and summarize the sensors that provide the input to SLAM.
We then introduce the RL methods used in this literature, from value-based and policy-based methods to actor-critic and deep RL.
Next, we classify RL applications in SLAM into three categories: path planning, including environment exploration and obstacle avoidance; loop closure detection; and active SLAM.
Thirteen representative studies are compared in terms of their simulation environment, deep learning method, SLAM method, and RL algorithm.
Most of these studies are evaluated mainly in simulation, and value-based methods from the deep Q-network family are the most common.
Finally, we discuss the challenges of applying RL to SLAM, namely computational demands, safety, generalization, high-dimensional state and action spaces, sample efficiency, and sensor and actuator delays, and we outline directions for future research.
\end{abstract}


\section{Introduction}
\label{sec:introduction}

Mobile robotics involves designing, constructing, operating, and utilizing robots to perform tasks in dynamic, non-fixed environments.
These robots are usually designed to be mobile and autonomous, capable of operating without direct human control~\cite{sharma2023review}.
Autonomous driving is one specific application of mobile robotics; it develops vehicles that navigate and operate on their own in real-world environments, such as roads and highways.
These vehicles combine sensors, such as cameras, radar, and Light Detection and Ranging (LiDAR), with algorithms that perceive their surroundings and decide how to navigate safely to their destination~\cite{othman2021public}.

Simultaneous Localization and Mapping (SLAM) is a key technology in Mobile Robotics and Autonomous Driving (MRAD)~\cite{bresson2017simultaneous}.
It enables a robot to build a map of an unknown environment while navigating, by continuously observing map features to determine its own position and orientation~\cite{ahmed2023active,ren2023slam}.
Localization determines the robot's position within the environment; mapping constructs a representation of the environment (the map) as the robot explores it~\cite{placed2023survey}.
Accurate localization against a map of the surroundings underpins tasks ranging from spatial exploration to autonomous driving, because it provides the information needed to predict obstacle movements and determine optimal maneuvers~\cite{wu2023research,taheri2021slam}.

Achieving accurate localization is difficult, however, because sensor measurements are inherently noisy.
A localization system must also handle outliers, occlusions, and sensor failures, and resolve scale differences between the map and the robot's motion~\cite{ahmed2023active}.
Detecting revisited locations, known as loop closures, is further complicated by perceptual aliasing and sensor limitations.
The estimate of the robot's pose must remain consistent over time, especially in dynamic environments~\cite{placed2023survey,zeng2020view}.
Finally, localization must run in real time to support effective navigation~\cite{diginsa2023review}.

Mapping raises further challenges for accuracy and reliability.
The first is aligning sensor measurements precisely with map features~\cite{bresson2017simultaneous,eyvazpour2023hardware}.
Constructing an accurate map also requires extracting meaningful features from sensor data, such as point clouds and images~\cite{li2022research}.
In addition, vehicle motion must be estimated accurately during mapping, and revisited locations must be identified to close loops in the map~\cite{liu2024particle}.
The map representation, whether a 2D grid, a 3D point cloud, or another format, must suit the specific mapping needs.
Finally, the map must remain consistent as new data are incorporated, so that it stays accurate and reliable over time~\cite{placed2023survey}.

These challenges drive research on SLAM algorithms that are more accurate, robust, and efficient.
Artificial Intelligence (AI) algorithms are integral to SLAM, as they enable robots to navigate and map their environment in real time.
In particular, Reinforcement Learning (RL) offers promising opportunities to improve exploration, localization, and map building in SLAM applications~\cite{fan2024integrating}.
In MRAD, RL aims to give robots and vehicles a systematic way to acquire, from sensor data, the skills needed to navigate environments and make informed decisions~\cite{botteghi2020reinforcement,hamza2022deep}.
This survey therefore reviews the practical uses of RL in SLAM.
Its main contributions are as follows:

\begin{itemize}
\item We present SLAM approaches in two distinct categories, passive and active SLAM.
  This classification gives a clearer framework for evaluating and comparing SLAM methods.
\item We examine the sensors that provide the input data for SLAM algorithms.
\item We classify RL applications in SLAM into three main categories: path planning, with environment exploration and obstacle avoidance as its two subcategories; loop closure detection; and active SLAM.
  This classification helps researchers explore the applications of RL in SLAM and develop new ideas for improving them.
\end{itemize}

The rest of the paper is organized as follows.
Section~\ref{sec:slam} introduces the types of SLAM approaches.
Section~\ref{sec:data_source} describes the sensors that capture information about the surrounding environment.
Section~\ref{sec:reinforcement_learning} explains the principles of RL.
Section~\ref{sec:classification} reviews and categorizes studies that have used RL in SLAM.
Section~\ref{sec:challenges} discusses the challenges of applying RL to SLAM.
Finally, Section~\ref{sec:future_directions} outlines future directions, and Section~\ref{sec:conclusion} concludes the paper.

\section{Simultaneous Localization and Mapping}
\label{sec:slam}

The probabilistic SLAM problem originated at the 1986 IEEE Robotics and Automation Conference in San Francisco~\cite{durrantwhyte2006simultaneous}.
In SLAM, a robot starting at an unknown location in an unknown environment incrementally builds a map of its surroundings and, at the same time, uses this map to estimate its own location~\cite{durrantwhyte2006simultaneous}.
SLAM thus denotes a family of approaches that let robots autonomously determine their location and map the surrounding environment as they move through it.
It combines two coupled components: (1) localization, which estimates the robot's pose relative to the map, and (2) mapping, which reconstructs the environment from on-board sensors such as cameras, visual--inertial units, and laser scanners~\cite{ahmed2023active,saeedi2011multiple}.
Modern SLAM systems commonly formulate the problem as a graph whose nodes are robot poses and landmark positions and whose edges encode the relative measurements between them~\cite{grisetti2010tutorial}.
Let $\mathbf{x}$ stack all these variables, where each planar robot pose consists of a 2D position and a heading angle, that is, an element of $SE(2)$.
Graph-based SLAM seeks the configuration $\mathbf{x}^{*}$ that minimizes the measurement errors $\mathbf{e}_{k}(\mathbf{x})$ of all constraints $k$, each weighted by its information matrix $\boldsymbol{\Omega}_{k}$ (the inverse of the measurement covariance), as expressed in \eqref{eq:1}~\cite{grisetti2010tutorial,ahmed2023active}.

\begin{equation}
\mathbf{x}^{*}=\arg\min_{\mathbf{x}}\sum_{k}\mathbf{e}_{k}(\mathbf{x})^{\top}\boldsymbol{\Omega}_{k}\,\mathbf{e}_{k}(\mathbf{x})
\label{eq:1}
\end{equation}

SLAM approaches can be divided into passive and active SLAM~\cite{placed2023survey}; the following subsections discuss each.

\subsection{Passive SLAM Approach}

Passive SLAM systems do not navigate the robot to explore unfamiliar environments.
Instead, the robot is controlled manually or follows predetermined waypoints~\cite{ahmed2023active}, so the SLAM system does not decide where it moves.
Passive SLAM is particularly suitable for scenarios where precise robot motion is not critical, and it focuses instead on effective mapping and localization~\cite{thrun2008simultaneous}.
This approach allows more predictable and controlled movement, which can be advantageous in specific applications.
However, it also implies that the robot does not choose its own motion to reduce the uncertainty of its map and pose; active SLAM provides this capability~\cite{ahmed2023active}.
Passive SLAM thus separates the control of robot motion from map and pose estimation.
It relies on the estimation back end alone, such as filtering or graph optimization, to estimate robot poses and build maps; a common filtering approach is the Particle Filter (PF).
A PF represents the posterior distribution with a set of particles (samples), each of which represents a potential robot pose and map hypothesis~\cite{zheng2023simultaneous}.

\subsection{Active SLAM Approach}

Active SLAM adds decision making to SLAM: the robot controls its own motion so as to reduce the uncertainty of its localization and map while exploring~\cite{placed2023survey}.
Active SLAM systems take sensor readings as input and make real-time decisions that influence future measurements~\cite{carrillo2012comparison,burgard1997active}.
Such a system typically follows a three-step process~\cite{makarenko2002experiment}:

{\renewcommand{\labelenumi}{\arabic{enumi}.}
\begin{enumerate}
\item identifying all potential locations for exploration (ideally infinitely many);
\item estimating the benefit of the actions that would move the robot from its present position to each of these locations; and
\item choosing and executing the most advantageous action.
\end{enumerate}}

Active SLAM includes modules for planning waypoints and generating trajectories.
It draws on information theory, optimal control theory, and RL to steer the robot toward its destination~\cite{ahmed2023active}.
Section~\ref{sec:classification} discusses the application of RL to active SLAM.

\section{Data Source}
\label{sec:data_source}

SLAM in autonomous driving typically integrates data from multiple sensors to perceive the surroundings comprehensively~\cite{lin2020orb,kumar2024advances}.
These sensors include wheel odometry, LiDAR, cameras, a Global Navigation Satellite System/Inertial Navigation System (GNSS/INS), an Inertial Measurement Unit (IMU), and radar.
Combining their measurements provides a detailed and accurate understanding of the environment~\cite{zheng2023simultaneous}.

Besides sensor data, the control commands sent to the vehicle's actuators, such as steering, throttle, and brakes, affect the robot's movement.
RL techniques can optimize these control maneuvers to improve SLAM performance~\cite{ahmed2023active,zheng2023simultaneous}.
SLAM algorithms usually start with an initial estimate of the robot's pose, derived from sources such as GNSS or wheel odometry~\cite{reitbauer2023liwo}.
RL approaches can also refine the pose estimate itself; for example, an RL agent trained for planar pose-graph optimization outperformed the g2o solver on challenging instances~\cite{kourtzanidis2023rlpgo}.

\emph{Wheel odometry:} Wheel odometry is a basic localization technique that measures the rotation of a robot's wheels with encoders to estimate its incremental movement in distance and direction~\cite{quan2019tightly}.
This method is computationally inexpensive and widely available, and it provides continuous pose updates as the robot moves~\cite{yousif2015overview}.
However, these incremental estimates accumulate errors over time because of wheel slippage and uneven terrain.
Encoder readings are also subject to noise, which reduces accuracy~\cite{mahmoud2023dynamic}.
In SLAM systems, wheel odometry is commonly used as one of the sensor inputs~\cite{quan2019tightly}.
RL algorithms can learn to integrate wheel odometry information for pose estimation.
For example, a Proximal Policy Optimization (PPO) agent (see Section~\ref{sec:reinforcement_learning}) learned to weight the pose estimates of LiDAR scan matching, initialized with wheel-encoder odometry, and of RGB-D visual SLAM, improving localization accuracy and stability~\cite{wong2024multi}.

\emph{LiDAR:} LiDAR, a remote sensing technology, uses laser light to measure distances and create detailed 3D maps of environments~\cite{li2020lidar}.
By emitting laser pulses and timing their reflections from objects, LiDAR produces accurate point clouds.
These point clouds are used to build comprehensive maps, detect obstacles, and plan collision-free paths~\cite{debeunne2020review}.
LiDAR also supports precise localization by matching current scans against previously mapped features.
LiDAR scans are thus central to both mapping and obstacle detection in autonomous navigation systems~\cite{tang2015lidar}.

\emph{Cameras:} Cameras capture images of the surroundings, from which visual features such as key points and edges are extracted to support localization and map construction~\cite{yang2020multi}.
These images are the basis of visual odometry, which estimates motion by tracking features across consecutive frames.
Cameras also help recognize previously visited locations and build detailed 3D maps~\cite{lang2024gaussian}.

\emph{GNSS/INS:} GNSS provides global position information, namely latitude, longitude, and altitude, from satellite signals.
INS, in contrast, estimates position and orientation from the accelerations and angular rates measured by accelerometers and gyroscopes~\cite{zheng2023simultaneous}.
GNSS therefore delivers an initial position estimate, and INS provides continuous tracking that bridges GNSS signal outages, maintaining accurate navigation even in difficult conditions~\cite{chang2020gnss}.

\emph{IMU:} IMUs combine accelerometers and gyroscopes to measure linear acceleration and angular velocity.
They provide short-term motion estimates, and their data are fused with those of other sensors, such as LiDAR and cameras, to make SLAM systems more robust~\cite{abdelaziz2023ins}.

\emph{Radar:} Radar sensors detect objects using radio waves, which makes them particularly valuable for obstacle detection in adverse weather.
In SLAM systems, radar improves accuracy, especially under poor lighting or occlusion~\cite{woo2018localization}.
The number of radar sensors in a SLAM system depends on the application, its requirements, and the desired accuracy~\cite{khan2022investigation}.
Some systems use a single radar for both motion estimation and mapping.
This configuration is particularly beneficial in low light or adverse weather, where cameras and LiDAR may be limited.
More advanced systems may use multiple radars to tolerate sensor failures and cover a full 360-degree field of view~\cite{hong2021radar}.
Combining data from multiple radars improves accuracy and reduces blind spots~\cite{huang2021high}.
Radar is also often fused with other sensors that provide complementary information~\cite{zhu2023camera}:

\begin{itemize}
\item LiDAR, which offers long-range detection and high-resolution mapping;
\item cameras, since radar--camera fusion helps in difficult lighting conditions; and
\item IMUs, which improve motion estimation.
\end{itemize}

Ongoing research seeks optimal configurations for radar-based SLAM that balance cost, power consumption, and sensor placement~\cite{lu2022low}.

\section{Reinforcement Learning}
\label{sec:reinforcement_learning}

A solid understanding of RL is needed to analyze data and develop intelligent automated applications~\cite{sarker2021machine}.
In RL, an agent learns by interacting with its environment and receiving rewards~\cite{sutton2018reinforcement}.
The agent explores various actions to determine which ones yield the highest rewards over time.
Because actions can have lasting effects, the return $R$ is computed by summing the discounted rewards over the time steps of an episode, as shown in \eqref{eq:2}, where $\gamma\in[0,1]$ denotes the discount factor, $T$ is the horizon, and $r(s_{t},a_{t})$ is the reward for action $a_{t}$ in state $s_{t}$~\cite{sutton2018reinforcement}.

\begin{equation}
R = \sum_{t=0}^{T} \gamma^{t}r(s_{t},a_{t})
\label{eq:2}
\end{equation}

A policy $\pi$ selects action $a$ in state $s$ with probability $\pi(a\mid s)$.
The value $V^{\pi}(s)$ of a state is the expected return obtained by starting in $s$ and following $\pi$, and it satisfies the Bellman equation \eqref{eq:3}~\cite{sutton2018reinforcement}.

\begin{equation}
\begin{aligned}
V^{\pi}(s) &= \mathbb{E}_{\pi}\left[R \mid s_{0}=s\right] \\
&= \mathbb{E}_{\pi}\left[r(s_{t},a_{t})+\gamma V^{\pi}(s_{t+1}) \mid s_{t}=s\right]
\end{aligned}
\label{eq:3}
\end{equation}

Here, $\mathbb{E}_{\pi}$ denotes the expectation over the trajectories generated by $\pi$.
Note that $r$ is the immediate reward, whereas $V^{\pi}$ accumulates the expected discounted rewards from $s$ onward~\cite{sutton2018reinforcement}.
Similarly, the action value $Q^{\pi}(s,a)$ of taking action $a$ in state $s$ and following $\pi$ afterwards satisfies the Bellman equation \eqref{eq:4}~\cite{arulkumaran2017brief}.

\begin{equation}
\begin{aligned}
Q^{\pi}(s,a) &= \mathbb{E}_{\pi}\left[R \mid s_{0}=s,a_{0}=a\right] \\
&= \mathbb{E}_{\pi}\big[r(s_{t},a_{t}) \\
&\qquad +\gamma Q^{\pi}(s_{t+1},a_{t+1}) \mid s_{t}=s,a_{t}=a\big]
\end{aligned}
\label{eq:4}
\end{equation}

In other words, the Q-value measures how good it is to commit to a particular first action and then behave according to $\pi$.
Averaging the action values over the actions that $\pi$ chooses gives the state value, as in \eqref{eq:5}, where $\mathcal{A}$ is the set of actions.

\begin{equation}
V^{\pi}(s)=\sum_{a\in \mathcal{A}} \pi(a\mid s)\,Q^{\pi}(s,a)
\label{eq:5}
\end{equation}

When the Markov Decision Process (MDP) has finite sets of states and actions, the action values can be stored in a Q-table.
The agent then takes actions and updates the Q-values in the table based on the rewards received from the environment.
In Q-learning, each observed transition $(s_{t},a_{t},r_{t},s_{t+1})$, with $r_{t}=r(s_{t},a_{t})$, updates the table according to \eqref{eq:6}, where $\alpha\in(0,1]$ is the learning rate~\cite{watkins1992q,singh2023review}.

\begin{equation}
\begin{aligned}
Q(s_{t},a_{t}) \leftarrow{} & Q(s_{t},a_{t}) \\
&+ \alpha\Big[r_{t}+\gamma\max_{a'}Q(s_{t+1},a')-Q(s_{t},a_{t})\Big]
\end{aligned}
\label{eq:6}
\end{equation}

\begin{figure}[!t]
\centering
\includegraphics[width=\columnwidth]{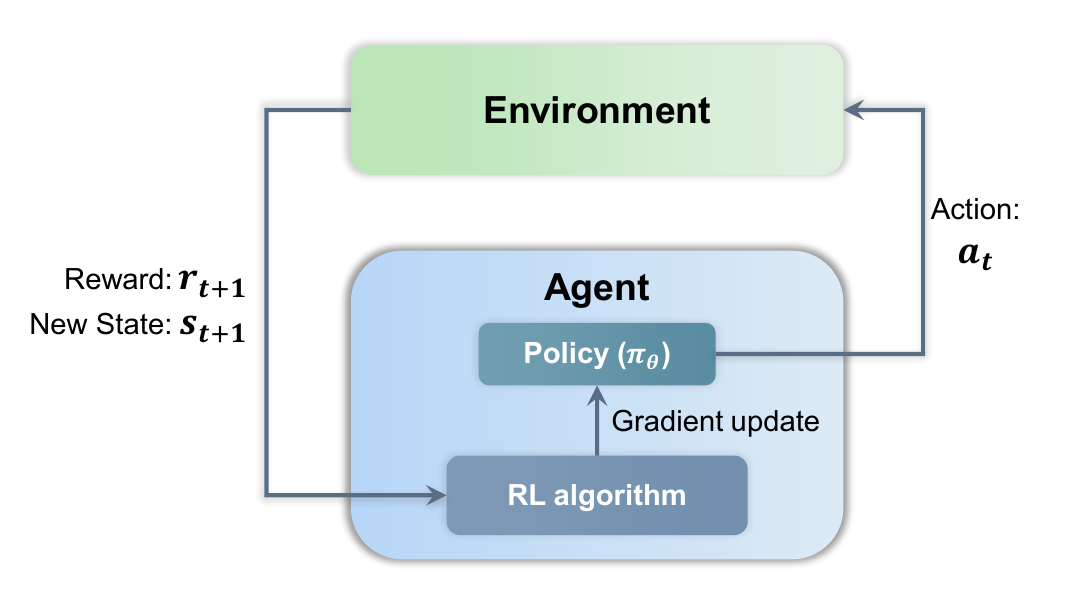}
\caption{The agent--environment interaction loop in reinforcement learning.}
\label{fig:fig1_rl}
\end{figure}

Fig.~\ref{fig:fig1_rl} illustrates an agent that learns from its environment using an RL algorithm.
The agent acts according to its policy, and the environment returns a reward and the next state.
The RL algorithm uses this feedback to update the policy and thereby improve the selection of actions.

RL methods fall into two primary families: value-based and policy-based methods.
In practice, the choice between them depends on the specifics of the problem, such as the nature of the state and action spaces, stability concerns, and the desired convergence properties~\cite{sewak2022policy}.

\subsection{Value-Based Methods}

Value-based methods estimate the value function, that is, the expected return that an agent can achieve from a given state (or state-action pair) while following a particular policy.
Their goal is to find an optimal value function from which an optimal policy can be derived~\cite{byeon2023advances}.
Common value-based methods include the following:

\begin{itemize}
\item Q-learning: This algorithm learns the action-value function (Q-function), which estimates the expected return of taking an action in a specific state and following the best policy thereafter~\cite{watkins1992q}.
  It is off-policy: it learns the values of the greedy policy while the agent follows an exploratory one.
  Once the Q-function is learned, the agent acts by selecting the actions that maximize it~\cite{clifton2020q}.
\item Deep Q-networks (DQN): This extension of Q-learning approximates the Q-function with a deep neural network, which makes it feasible to scale Q-learning to environments with high-dimensional state spaces~\cite{mnih2015human,huang2020deep}.
\item SARSA: State-Action-Reward-State-Action (SARSA) is the on-policy counterpart of Q-learning~\cite{rummery1994online}: it updates the Q-values based on the action actually taken by the agent, so the learned values reflect the policy being followed.
  Its name comes from the quintuple of events used in each update, $(s_{t},a_{t},r_{t},s_{t+1},a_{t+1})$.
  In each step, after the agent takes an action and receives a reward, SARSA updates its Q-value estimate by considering the next state and the next action that the policy selects~\cite{sutton2018reinforcement}.
\end{itemize}

\subsection{Policy-Based Methods}

Policy-based methods learn a policy directly.
The policy can be either deterministic or stochastic, and it maps states (or observations) directly to actions, without necessarily using a value function to do so.
The objective is typically to optimize the policy itself so that it maximizes a measure of cumulative reward~\cite{sewak2022policy}.
The main policy-based methods are the following.

\emph{Policy gradient methods:} These methods optimize the policy by gradient ascent.
They compute the gradient of the expected return with respect to the policy parameters and use it to update the policy in a direction that improves performance~\cite{sutton2000policy}.

\begin{itemize}
\item REINFORCE: This foundational policy gradient algorithm was proposed by Williams (1992).
  It estimates the policy gradient by Monte Carlo sampling; it is simple but often suffers from high variance, which slows convergence in some scenarios.
  REINFORCE directly optimizes the expected return of the policy by adjusting the weights of the policy network along sampled trajectories~\cite{williams1992simple}.
\item Trust Region Policy Optimization (TRPO): TRPO extends the basic policy gradient method with a trust region constraint.
  The constraint keeps each updated policy close to the current one, with the difference measured by the Kullback--Leibler (KL) divergence.
  Enforcing this constraint stabilizes training.
  However, this stability comes at the expense of increased computational complexity~\cite{schulman2015trust}.
  Because each update is kept small, TRPO improves the policy incrementally, which enhances both stability and performance.
  TRPO has also been used to train mobile-robot navigation policies~\cite{tai2018socially}.
\item Proximal Policy Optimization (PPO): PPO is a refinement of TRPO that addresses its computational inefficiency.
  It replaces the trust region constraint with a simpler clipping mechanism in the surrogate loss function, which limits large policy updates.
  PPO is computationally efficient, easy to implement, and performs well across a wide range of tasks, which has made it one of the most popular policy gradient algorithms~\cite{schulman2017proximal}.
  It has been applied to active SLAM, for example to train the global exploration policy of Active Neural SLAM~\cite{chaplot2020learning}.
\item Actor-critic methods: These methods combine the advantages of value-based and policy-based methods.
  The actor selects actions according to the current policy, and the critic evaluates them by estimating the value function; this feedback loop continually refines both the policy and the value estimates~\cite{grondman2012survey}.
  The combination allows efficient learning and convergence as well as quick adaptation to changes in the environment, which real-time autonomous navigation and mapping require~\cite{grondman2011efficient}.
\end{itemize}

\subsection{Deep Reinforcement Learning}

Deep Reinforcement Learning (DRL) is a machine learning approach that combines RL with deep neural networks~\cite{wang2022deep}.
Unlike traditional methods, DRL learns feature representations directly from raw data, such as images and sensor readings, instead of relying on handcrafted features~\cite{zai2020deep}.
It has shown promise in solving complex decision-making problems and has been applied in fields such as robotics, gaming, and autonomous systems~\cite{singh2023review,lei2018dynamic}.

Value-based DRL methods, such as DQN and its extensions, have proven effective for decision-making problems with discrete and low-dimensional action spaces.
However, these methods face significant limitations in tasks with continuous, high-dimensional action spaces.
Active SLAM, which involves complex path planning and precise control, is an example of such a task~\cite{zhao2024exploration}.
To overcome these limitations, researchers have increasingly turned to policy gradient and actor-critic methods.
These methods handle continuous, high-dimensional action spaces more effectively, which enables robots to navigate and map their environments autonomously.
This shift reflects the growing role of advanced RL techniques in active SLAM~\cite{singh2023review}.

\subsection{RL-Based SLAM Applications}

RL has had a significant impact on practical SLAM applications, particularly in robotics, autonomous vehicles, and drones.
Traditionally, SLAM has struggled in complex and dynamic environments, where sensor noise, data association, and computational constraints can degrade performance.

\emph{Robotics:} RL-based SLAM enables robots to navigate and map complex environments autonomously by learning optimal navigation strategies through trial and error.
This approach improves the robots' ability to make real-time decisions and to adapt to dynamic conditions in unfamiliar indoor settings while avoiding obstacles~\cite{singh2023review}.
Whereas traditional systems cannot learn on their own, Lee et al.~\cite{lee2022mobile} introduced an end-to-end DRL approach for autonomous navigation in uncharted environments.
They developed two deep Q-learning agents, a DQN and a double DQN, that allow robots to learn collision avoidance and navigation skills independently.
Their pipeline first detects the target object with a deep neural network and then guides navigation with deep Q-learning.
More recent research has focused on improving SLAM for mobile robots in complex environments.
Wong et al.~\cite{wong2024multi} developed a multi-sensor fusion approach based on DRL and Multi-Model Adaptive Estimation (MMAE), which improved the localization precision and stability of robots in challenging conditions.
Su et al.~\cite{su2024aslam} presented ASLAM-FD, a framework that combines adaptive multi-sensor fusion with degradation detection and DRL to maintain SLAM accuracy in dynamic environments.
Liu et al.~\cite{liu2024active} developed a DRL-based active SLAM system with multi-sensor fusion that enables snake robots to navigate autonomously and avoid obstacles in hard-to-access areas.
The system, which uses a lightweight structure with a 2D LiDAR and an IMU, significantly improved path planning and reduced collision rates compared with traditional methods.

\emph{Autonomous vehicles:} RL methods have become essential for real-time SLAM tasks in autonomous cars, particularly those operating in dynamic environments such as urban streets.
These vehicles use RL for decision-making tasks such as lane keeping~\cite{wang2020lane}, obstacle avoidance~\cite{arvind2020autonomous}, and route planning~\cite{rezaee2021motion}, while constructing and updating maps of their surroundings.

RL excels in such applications because it can handle high-dimensional sensory data (e.g., from LiDAR and cameras) and learn optimal control policies through interaction with the environment.
For example, RL enables autonomous cars to adapt to dynamic traffic conditions, manage sensor noise, and make split-second decisions that ensure safety and efficiency~\cite{tang2024deep}.
Advanced RL techniques, such as actor-critic methods and PPO, further improve stability and convergence during training.
These methods allow vehicles to navigate effectively in complex scenarios, such as unstructured roads, crowded intersections, and multi-agent environments, where traditional SLAM approaches might struggle~\cite{song2023reaching}.

\begin{figure*}[!t]
\centering
\includegraphics[width=0.85\textwidth]{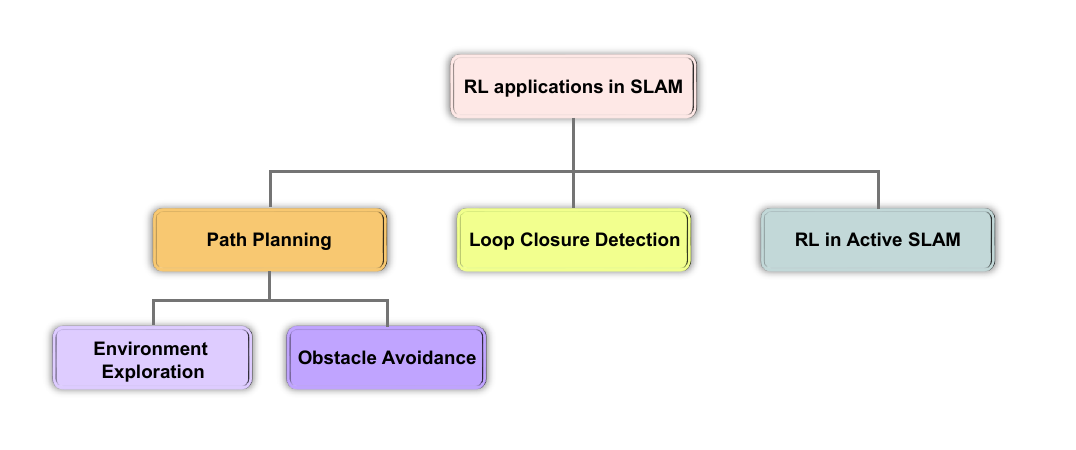}
\caption{Classification of RL applications in SLAM.}
\label{fig:fig2_taxonomy}
\end{figure*}

\emph{Drones:} RL is particularly useful for drones, which rely on SLAM in GPS-denied environments.
For example, RL has enabled drones to race at high speed, processing rapid sensor inputs and adjusting their trajectories dynamically.
In real-world experiments, RL-based controllers outperformed classical optimal control methods, especially in the presence of unmodeled dynamics or disturbances~\cite{tang2024deep,song2023reaching}.

\section{Classification of Existing Methods}
\label{sec:classification}

RL has many applications within SLAM.
This section reviews them and classifies them into three categories, as shown in Fig.~\ref{fig:fig2_taxonomy}: path planning, with environment exploration and obstacle avoidance as its subcategories; loop closure detection; and active SLAM.

\subsection{Path Planning}

Robot path planning optimizes criteria such as work cost and route length, so that the robot navigates efficiently while avoiding obstacles~\cite{baziyad2021addressing}.
Path planning methods can generally be categorized into traditional and AI approaches.
Depending on the available environmental information, they can be further classified into global planning, in a known environment, and local planning, in an unknown one~\cite{chen2024deep}.

Wang~\cite{ruiqi2024research} developed a visual SLAM system, based on the ORB-SLAM3 framework, that generates a dense point cloud map.
This map is converted into an OctoMap and then projected onto a 2D grid map, on which an RL-based path planning algorithm operates.
In experimental comparisons of Q-learning, DQN, and SARSA, DQN converged fastest and performed best, particularly in complex, high-dimensional environments.

Nam et al.~\cite{nam2023slam} introduced a navigation framework for mobile robots that integrates SLAM and DRL to improve operational efficiency.
SLAM builds the map and localizes the robot, and an Ant Colony Optimization (ACO) algorithm plans a predetermined route.
Where obstacles in the environment change, the framework uses DRL for local path planning.
The authors also compared three DRL-based navigation algorithms: Deep Q-Network (DQN), Twin Delayed Deep Deterministic Policy Gradient (TD3), and Proximal Policy Optimization (PPO).

\subsubsection{Environment Exploration}

Robot exploration combines exploration methods and technologies that enable robots to navigate, map, and explore unfamiliar environments autonomously and efficiently, drawing on advances in robotics, AI, and sensor technology~\cite{chen2024deep}.

Chen et al.~\cite{chen2024deep} introduced a DRL-based exploration model that navigates unknown environments without collisions.
It combines SLAM with a dual-mode DRL structure to escape local minima.
After 300,000 training rounds, the model achieved zero collisions and almost no repeated exploration, and its explored area was within 5\% of that of the ROS slam-gmapping method.

Li et al.~\cite{li2020deep} studied automatic exploration of unfamiliar environments using DRL together with Karto SLAM, a graph-based SLAM technique.
Their framework consists of decision, planning, and mapping modules; the decision module uses a deep neural network to learn exploration strategies from the partial map.

\subsubsection{Obstacle Avoidance}

Obstacle avoidance involves algorithms and technologies that enable vehicles to perceive obstacles in their surroundings accurately and in real time and to plan collision-free motion around them, ensuring safe navigation~\cite{iqbal2020obstacle}.

Wen et al.~\cite{wen2020path} used fully convolutional residual networks to estimate depth images for obstacle recognition.
The Dueling DQN algorithm plans the obstacle-avoidance path, and FastSLAM builds a two-dimensional map of the environment.

According to Nam et al.~\cite{nam2023inav}, SLAM algorithms are effective for mapping but struggle to avoid dynamic obstacles, whereas DRL algorithms can handle dynamic situations but may be sub-optimal when used alone.
Their method therefore combines the two: SLAM fuses data from several sensors, including LiDAR, to map the environment; ACO plans the shortest global path; and DRL plans the local path, so that the robot can make decisions based on its current situation and goal.

Fayjie et al.~\cite{fayjie2018driverless} used DRL for autonomous navigation and obstacle avoidance in self-driving cars.
The study fed camera and laser-sensor data to a neural network trained with DQN and tested it in autonomous-driving simulations.

\subsection{Loop Closure Detection}

Loop closure addresses inaccuracies in the sensor measurements of robots and autonomous vehicles~\cite{arshad2021role}.
It also mitigates other effects that reduce reliability, such as drift, acceleration changes, and weather conditions.
By detecting when a vehicle revisits a previously visited location, loop closure corrects the errors accumulated in the map and the position estimate~\cite{tsintotas2022revisiting}.
It is therefore key to the accuracy and reliability of the vehicle's navigation system and to its overall performance~\cite{charroud2024localization}.

Iqbal et al.~\cite{iqbal2022deep} treated loop closure detection as a reward-driven optimization problem solved with DRL.
A policy was trained in a grid-based environment that simulates indoor spaces, real-world maps and features were converted into this representation, and the agent was tested in simulation and in indoor lab environments.

In his Ph.D. dissertation, Iqbal~\cite{iqbal2019classified} presented two approaches: a non-parametric statistical association and clustering method that localizes classified objects in visual SLAM, and a DRL approach that treats loop closure detection as a reward-driven optimization process.
The DRL approach was implemented in a simulated grid environment, where the agent learned from the generated data to detect loop closures in different environments.

\subsection{RL in Active SLAM}

As described in Section~\ref{sec:slam}, active SLAM lets robots and autonomous systems explore their surroundings actively.
The system continuously improves its understanding of the environment while updating its position in real time~\cite{placed2020deep}.
By deciding where to move next, the robot gathers the most valuable data for mapping and localization.
Active SLAM thus lets the system adapt to changing environments and navigate efficiently through unknown areas~\cite{chen2020active,palomeras2019active}.

Fang et al.~\cite{fangslamuzero} used MuZero to improve agents' planning in joint active SLAM and navigation, in which an agent navigates an unfamiliar environment while mapping it and localizing itself.
Their SLAMuZero framework combines SLAM with the tree-search-based MuZero algorithm: an explicit encoder-decoder architecture builds the map, and a prediction function estimates the policy and value from this map.
SLAMuZero was trained on 1 million frames instead of the 10 million used by baselines and reached a coverage of 0.961, compared with 0.948 for Active Neural SLAM~\cite{chaplot2020learning}.

Placed and Castellanos~\cite{placed2020deep} used deep Q-learning with laser measurements for navigation, with the aim of reducing the uncertainty of the robot's localization and map.
Their trained agents reduced uncertainty, transferred knowledge to new maps, and learned to navigate and explore in simulation.

Pei et al.~\cite{pei2020active} introduced active relative localization for multi-agent SLAM.
Their DRL-based task allocation algorithm uses a Multi-Agent System DQN (MAS-DQN) to improve collaboration efficiency in SLAM.

Alcalde et al.~\cite{alcalde2022da} used two agents, a completeness-based and an uncertainty-based agent.
Both agents completed maps without collisions; the uncertainty-based agent generated longer paths but better maps, and both performed comparably to or better than frontier-based and active loop-closing baselines.

Table~\ref{tab:comparison} compares the reviewed studies in terms of their simulation environment, deep learning method, SLAM method, and RL algorithm, together with their advantages and disadvantages.
Most of these studies are evaluated mainly in simulation, with Gazebo the most common simulator, and nine of the thirteen use value-based methods from the DQN family.

\begin{table*}[!tp]
\caption{Comparison of Reviewed Studies.}
\label{tab:comparison}
\centering
\footnotesize
\setlength{\tabcolsep}{3pt}
\renewcommand{\arraystretch}{1.05}
\newcommand{\tabitem}{\par\noindent\hangindent=0.8em\hangafter=1\makebox[0.8em][l]{\textbullet}}
\newcommand{\category}[1]{\multicolumn{8}{@{}l}{\textbf{\textit{#1}}}\\[1pt]}
\begin{tabularx}{\textwidth}{@{}P{1.72cm}P{0.6cm}P{2.1cm}P{1.55cm}P{1.85cm}P{1.75cm}LL@{}}
\toprule
\textbf{Study} & \textbf{Year} & \textbf{Simulation environment} & \textbf{DL method}
  & \textbf{SLAM method} & \textbf{RL algorithm} & \textbf{Advantages} & \textbf{Disadvantages} \\
\midrule
\category{Path planning}
Wang~\cite{ruiqi2024research} & 2024 & 2D grid map (RL); TUM RGB-D and recorded maze (SLAM)
  & Deep neural network & ORB-SLAM3 & Q-learning, DQN, SARSA
  & \tabitem Effective map conversion \tabitem Contribution to autonomous navigation
  & \tabitem Global planning only \tabitem Discrete actions \tabitem Simulation only \\ \addlinespace[3pt]
Nam et~al. \cite{nam2023slam} & 2023 & Gazebo, ROS~2, TurtleBot3
  & Deep neural network & SLAM + MCL & DQN, PPO, TD3
  & \tabitem Comprehensive framework \tabitem Comparison of three DRL algorithms
  & \tabitem Limited generalization \tabitem Potential overfitting \\
\midrule
\category{Environment exploration}
Chen et~al. \cite{chen2024deep} & 2024 & Gazebo, ROS
  & CNN & SLAM (package not named) & DQN
  & \tabitem Effective training strategy \tabitem Addresses real-world challenges
  & \tabitem Slower exploration \tabitem No real-world implementation \\ \addlinespace[3pt]
Li et~al. \cite{li2020deep} & 2020 & ROS Stage; real robot (RoboMaster)
  & CNN & Karto SLAM (GMapping in experiments) & DQN (dueling FCQN)
  & \tabitem Modular framework \tabitem Efficient exploration strategy \tabitem Generalization performance
  & \tabitem Discrete action space \\
\midrule
\category{Obstacle avoidance}
Nam et~al. \cite{nam2023inav} & 2023 & Gazebo, ROS~2 (DDS communication)
  & Deep neural network & SLAM-MCL & DQN, PPO, TD3
  & \tabitem Adaptability to dynamic environments \tabitem Exploration of multiple DRL algorithms
  & \tabitem Complexity in implementation \tabitem Restricted problem scope \\ \addlinespace[3pt]
Wen et~al. \cite{wen2020path} & 2020 & Gazebo
  & Fully convolutional residual network & FastSLAM & Dueling DQN
  & \tabitem Scalable action space \tabitem Improved learning efficiency
  & \tabitem High computational complexity \tabitem Sensor dependency \\ \addlinespace[3pt]
Fayjie et~al. \cite{fayjie2018driverless} & 2018 & Unity game engine
  & CNN & None & DQN
  & \tabitem Sensor fusion (LiDAR and camera)
  & \tabitem Simulation-limited validation \tabitem Simplistic action set \\
\midrule
\category{Loop closure detection}
Iqbal et~al. \cite{iqbal2022deep} & 2022 & Grid-based simulated indoor environment; indoor lab tests
  & CNN & Visual SLAM & DRL (MDP formulation)
  & \tabitem Real-world application \tabitem Improved feature utilization
  & \tabitem Dependence on prior knowledge \tabitem Computational demands \tabitem Hardware-dependent constraints \\ \addlinespace[3pt]
Iqbal~\cite{iqbal2019classified} & 2019 & Simulated grid environment
  & CNN & Visual SLAM & DRL (reward-driven)
  & \tabitem Robust data association method \tabitem Scalable to unknown environments
  & \tabitem Dependency on accurate depth estimation \tabitem Computational overheads \\
\midrule
\category{Active SLAM}
Fang et~al. \cite{fangslamuzero} & 2024 & Habitat (Gibson)
  & Encoder-decoder & Learned neural SLAM module & MuZero (model-based RL with MCTS)
  & \tabitem Efficiency in training \tabitem Improved performance
  & \tabitem High initial complexity \tabitem Unclear scalability \\ \addlinespace[3pt]
Placed and \mbox{Castellanos} \cite{placed2020deep} & 2020 & Gazebo
  & Deep neural network & GMapping (modified) & DQN, DDQN, D3QN
  & \tabitem Strong generalization capabilities \tabitem Complex simulation validation
  & \tabitem No real-world testing \tabitem Performance variability \\ \addlinespace[3pt]
Pei et~al. \cite{pei2020active} & 2020 & ROS simulation; ``telobot'' robots with Kinect
  & Deep neural network & ORB-SLAM & MAS-DQN (Rainbow-based)
  & \tabitem Attribute-aware task allocation \tabitem Realistic simulation
  & \tabitem Lack of a distributed alternative \tabitem Dependence on centralized coordination \\ \addlinespace[3pt]
Alcalde et~al. \cite{alcalde2022da} & 2022 & Gazebo, ROS, ROBOTIS TurtleBot3 Burger
  & Deep neural network & GMapping (lightweight passive SLAM back end) & PPO
  & \tabitem Focus on map quality and robustness \tabitem Efficiency through dimensionality reduction
    \tabitem Flexible and lightweight approach
  & \tabitem Limited physical-world validation \tabitem Dependency on reward function design \\
\bottomrule
\end{tabularx}

\vspace{3pt}
\parbox{\textwidth}{\scriptsize\raggedright
\textit{Abbreviations:} CNN, convolutional neural network; DL, deep learning; D3QN, dueling double DQN; DDQN, double DQN;
DDS, Data Distribution Service; DQN, deep Q-network; DRL, deep RL; FCQN, fully convolutional Q-network;
MAS-DQN, multi-agent system DQN; MCL, Monte Carlo localization; MCTS, Monte Carlo tree search;
MDP, Markov decision process; PPO, proximal policy optimization; RL, reinforcement learning;
SARSA, state-action-reward-state-action; TD3, twin delayed deep deterministic policy gradient.}
\end{table*}

\section{Challenges in Applying RL to SLAM}
\label{sec:challenges}

RL lets autonomous systems learn and adapt through interaction with their environment, but deploying it in real-world applications faces considerable challenges~\cite{zhao2024survey}.
These challenges must be overcome before RL can realize its full potential in SLAM; the main ones are the following.

\emph{Computational demands:} RL models, especially those built on deep learning, often require significant computational resources to perform well.
The need for high processing power and substantial memory can be a critical limitation when decisions must be made quickly~\cite{tourani2022visual}.
Achieving real-time performance with these models on edge devices, such as robots or drones, is therefore particularly difficult.
Such devices often have limited processing capabilities, which can prevent the use of sophisticated RL algorithms, even though these algorithms could enhance the devices' autonomy and operational effectiveness~\cite{favorskaya2023deep}.

\emph{Safety and reliability:} In RL-based SLAM, guaranteeing safe exploration and robust decision-making is critical, especially for autonomous vehicles navigating dynamic and complex urban environments.
Urban streets present many unpredictable conditions, such as varying traffic patterns, pedestrian movement, and environmental changes~\cite{bavle2023slam}.
These factors can impair a vehicle's ability to map and understand its surroundings accurately, which path planning and hazard avoidance depend on~\cite{tourani2022visual}.

\emph{Generalization:} RL models often struggle to adapt to diverse environmental conditions, because the controlled settings in which they are typically trained differ markedly from complex, unstructured real-world scenarios.
The assumptions and constraints built into simulation environments do not capture the full variability and unpredictability of natural settings.
As a result, RL models applied outside their training domains often perform sub-optimally, which reveals a critical gap in their ability to generalize~\cite{favorskaya2023deep}.

\emph{High-dimensional state and action spaces:} The state space contains all possible states of the environment, which in SLAM include the sensor data.
The action space contains the actions available to the agent~\cite{sutton2018reinforcement}; in SLAM, these actions are movements such as turning and accelerating~\cite{zhao2024exploration}.
SLAM systems process high-dimensional sensor data, from camera images to LiDAR point clouds, to navigate complex environments~\cite{kudriashov2020slam}.
However, the computational complexity of RL grows rapidly with the dimensionality of such spaces, which limits the effectiveness of RL agents~\cite{dulacarnold2021challenges}.

\emph{Sample efficiency:} Sample efficiency is the ability of an RL algorithm to learn from a small number of interactions (samples) with the environment~\cite{buckman2018sample}.
It matters because collecting real-world data for SLAM with autonomous vehicles, for example with laser sensors, can be costly and time-consuming~\cite{pang2024low}.

\emph{Sensor and actuator delays:} Sensor and actuator delays are the time gap between perceiving an event and responding to it.
This latency is inherent to the system: sensing, decision-making, and action execution each take time and must be precisely synchronized~\cite{hester2013texplore}.
RL algorithms must account for these delays to respond in a timely and precise manner and ensure smooth navigation and operation~\cite{hester2013texplore}.
In SLAM, such delays directly affect the timeliness of localization and mapping, so real-time processing is essential~\cite{dulacarnold2021challenges}.

\section{Future Directions}
\label{sec:future_directions}

Several areas are promising for future research on RL in SLAM:

\begin{itemize}
\item Adaptive sensor fusion: Combining data from sensors such as cameras, LiDAR, and IMUs is crucial for robust SLAM.
  Future work could develop RL agents that learn to fuse these modalities adaptively.
  Such agents could significantly improve the performance and reliability of SLAM systems, particularly in diverse and dynamic environments.
\item Self-supervised learning and data augmentation: Self-Supervised Learning (SSL) and data augmentation have substantial potential for improving RL-based SLAM, especially in MRAD applications.
  Exploiting large amounts of unlabeled data and generating diverse training samples can make SLAM systems more robust and improve their generalization.
  Future research should explore SSL strategies and data augmentation methods that make RL more effective in SLAM.
\item Knowledge transfer across environments: To be effective in real-world applications, RL agents must be able to transfer knowledge across different maps and environments.
  Future studies should investigate techniques such as domain adaptation and meta-learning to improve the generalization of RL-based SLAM systems.
  These techniques can let RL agents apply knowledge learned in one environment to another, which improves their adaptability and performance in previously unseen settings.
\end{itemize}

Progress in these areas would lead to more robust, efficient, and versatile RL-based SLAM systems and advance navigation in MRAD.

\section{Conclusion}
\label{sec:conclusion}

SLAM allows mobile robots and autonomous vehicles to map unknown environments while localizing themselves within them, and RL offers a way to learn the decisions that such systems must make.
In this survey, we reviewed the applications of RL in SLAM.
We distinguished passive from active SLAM, summarized the sensors that provide the input to SLAM, and introduced the value-based, policy-based, actor-critic, and deep RL methods used in this literature.
We then classified RL applications in SLAM into three categories: path planning, including environment exploration and obstacle avoidance; loop closure detection; and active SLAM.

The thirteen studies compared in Table~\ref{tab:comparison} show how RL is typically used in SLAM.
RL mostly serves as the decision layer of the system: it plans collision-free paths, chooses where to explore next, detects loop closures, and steers active SLAM toward motions that reduce the uncertainty of the map and the pose.
Most of these studies are evaluated mainly in simulation, with Gazebo the most common simulator.
Value-based methods from the DQN family dominate, although policy gradient and actor-critic methods such as PPO are increasingly used for continuous, high-dimensional action spaces.

This survey covers a representative rather than an exhaustive set of studies, and it focuses on mobile robots and autonomous vehicles.
Within this scope, RL-based SLAM still faces the challenges discussed in Section~\ref{sec:challenges}: high computational demands, safety and reliability requirements, limited generalization beyond the training environment, high-dimensional state and action spaces, low sample efficiency, and sensor and actuator delays.
Sensor configurations and the handling of environmental factors also still need to be tailored to each application domain.
Addressing these challenges, through RL agents that fuse sensors adaptively, self-supervised learning and data augmentation, and knowledge transfer across environments, while narrowing the gap between simulation and real-world deployment, is the key next step toward robust RL-based SLAM for MRAD.

\bibliographystyle{IEEEtran}
\bibliography{references}

\end{document}